\documentclass[acmsmall]{acmart}

\setcopyright{none}
\renewcommand\footnotetextcopyrightpermission[1]{} 
\begin{document}
\title{Capsule Networks for Protein Structure Classification and Prediction}

\author{Dan A. Rosa de Jesus}
\affiliation{%
  \institution{University of Puerto Rico, Lawrence Berkeley National Laboratory}
  \country{USA}}
  
\author{Julian Cuevas}
\affiliation{%
  \institution{University of Puerto Rico, Lawrence Berkeley National Laboratory}
  \country{USA}}

\author{Wilson Rivera}
\affiliation{%
  \institution{University of Puerto Rico, Lawrence Berkeley National Laboratory}
  \country{USA}}
  
  \author{Silvia Crivelli}
\affiliation{%
  \institution{Lawrence Berkeley National Laboratory}
  \country{USA}}
  
\renewcommand\shortauthors{D. Rosa, J. Cuevas, W. Rivera and S. Crivelli}

\begin{abstract}
Capsule Networks have great potential to tackle problems in structural biology because of their attention to hierarchical relationships. This paper describes the implementation and application of a Capsule Network architecture to the classification of {\it RAS} protein family structures on GPU-based computational resources. The proposed Capsule Network trained on 2D and 3D structural encodings can successfully classify {\it HRAS} and {\it KRAS} structures. The Capsule Network can also classify a protein-based dataset derived from a PSI-BLAST search on sequences of KRAS and HRAS mutations. Our results show an accuracy improvement compared to traditional convolutional networks, while improving interpretability through visualization of activation vectors. 

\end{abstract}

\ccsdesc[500]{Computing methodologies~Neural networks}
\ccsdesc[300]{Computing methodologies~Parallel programming languages}
\ccsdesc[500]{Applied computing~Molecular structural biology}

%
%

\keywords{deep learning, Capsule Networks, protein classification, protein structure prediction, interpretability.}

\maketitle
\thispagestyle{empty}
\section{Introduction}
Proteins are responsible for most functions in our body. They are made as an extended chain of amino acids and fold into a 3D structure that determines their function \cite{Dill2012-hv}. Determining their 3D structure is key to understanding how they work, why they cause diseases and how to design drugs to block or activate their functions \cite{Schaffhausen2012-km}. While experimental sequence generation is relatively cheap, it is challenging and expensive to classify and predict protein structure from sequences using experimental methods such as X-ray crystallography or NMR spectroscopy. Computational based prediction methods have the potential to reduce the burden cost of 3D protein structure analysis.  

Convolutional Neural Networks (CNNs) have been applied to structural biology. This approach enables computers to classify proteins or predict their structures by modeling the way the human brain processes inputs of information through different layers of representation. However, the computational cost associated with training CNNs increases when the networks are provided with a large number of data channels as required in complex protein structural problems. In addition CNNs do not take into account important spatial hierarchies between simple and complex objects which is also very important in protein structure classifications. Finally, it is hard to explain the rationale behind CNN models’ decisions. Thus, enhancing model's predictions with interpretability mechanisms is highly valuable.  

Recently, Hinton et al. \cite{Sabour2017} proposed Capsule Networks which introduce a new building block that can be used in deep learning to better model hierarchical relationships inside of internal knowledge representation of a neural network. This new development has the potential to help overcome the limitations of traditional CNNs when applied to protein structure problems.  

In this paper, we discuss the implementation and interpretability improvements of a Capsule Network applied to structural biology. The contributions of this paper are twofold: {\bf (1)} The implementation and application of a Capsule Network architecture to the classification of {\it RAS} protein family structures on GPU-based computational resources. The results show that the proposed Capsule Network trained on 2D and 3D structural encodings can successfully classify {\it RAS} family protein structures. The Capsule Network can also classify a protein-based dataset derived from a PSI-BLAST search on sequences of {\it RAS} mutations. {\bf (2)} The implementation of mechanisms for step by step interpretability of activation vectors and post-hoc interpretability of the Capsule Network. The ultimate goal is to demonstrate how the network ensembles pieces of knowledge to arrive at specific decisions and why those decisions are made. The internal activation of capsules is visualized and the information encoded in the Capsule Network is used for explanatory purposes.

This paper is organized as follows: In Section II, a review of related work on the application and interpretability of deep learning models in protein structure classification and prediction is presented. In Section III, the data representation and implementation details of the Capsule Network are discussed. Experimental results are presented in Section IV. Finally, conclusions and future work are discussed in Section V.

\section{Background and Related Work}

In this section, we discuss the related work in two specific areas: The application of deep learning in structural biology and the interpretability aspects of deep learning networks. 

\subsection{Deep Learning in Structural Biology}

Recently, the application of deep learning techniques in protein structure analysis has gained traction. For example, Zhou \cite{Zhou-2014} proposed a generative stochastic network (GSN) based method to predict local secondary structures. Spencer et al. \cite{Spencer2015-fk} proposed a protein structure predictor that uses deep learning network architectures combined with the position-specific scoring matrix generated by PSI-BLAST. They used a restricted Boltzmann machine based deep network. Wang et. al. \cite{Wang2016-ra}  proposed DeepCNF, a deep learning extension of Conditional Neural Fields (CNF). AtomNet \cite{Wallach2015-xc} is a deep convolutional neural network that can be applied to the analysis of bioactivity of small molecules for drug discovery applications. MoleculeNet \cite{Wu2018-vk} expanded on AtomNet by adding many more features to each voxel, including partial charge and atomic mass. CNNs have also been used to recognize protein-ligand interactions \cite{Ragoza2017-lf, Gomes2017-pp}.

In spite of all these advances, CNNs cannot preserve spatial relationship between components of an object because, some features will be discarded during the pooling process. CNNs compensate this deficiency by increasing the number of training data, a process known as data augmentation. Capsule Networks \cite{Sabour2017} have been proposed as a better approach to deep learning. A capsule is a group of neurons whose activation vector represents a specific type of entity. The length of the activation vector represents the probability that the entity exists and the orientation represents the instantiation parameters of the entity. Instead of using pooling to reduce dimensionality in CNNs, Capsule Networks implements a routing by agreement strategy in which outputs are sent to all parent capsules in the next layer. Each Capsule tries to predict the output of the parent capsules, and if this prediction conforms to the actual output of the parent capsule, the coupling coefficient between these two capsules increases. The work presented in this paper is a first approach to the application of capsule networks to structural biology. The results are promising and it opens a unique opportunity to explore issues related to interpretability of deep learning for computational structural biology.  

\subsection{Interpretability of Deep Learning Models}

Deep learning models do not provide information on their internal processing actions. while deep neural networks may generate more accurate predictions by learning nonlinear interactions between input variables, at the same time it makes explaining deep learning models very difficult. 

The are several interpretability works particularly in computer vision. Goferman et al. \cite{Goferman2012-es} used saliency maps to highlight pixels of the input image that are more relevant to the output classification. However, pixels identified as salient regions are not necessarily the pixels being involved in making the predictions. In addition salient maps do not explicitly deal with hidden layers. Ribeiro et al \cite{Ribeiro2016-rf} proposed Local Interpretable Model-agnostic Explanations (LIME), which explains the prediction by approximating the original model with an interpretable model around several local neighborhoods. The Gradient-weighted Class Activation Mapping (Grad-CAM) \cite{Selvaraju2016-if} approach is another explanation method for CNNs which uses gradient to obtain localization map as a visual explanation and finds important layers for each class. A summary of interpretability techniques is provided by Montaven et al. \cite{Montavon2017-zl}. They evaluated the performance of several recently proposed techniques of interpretation.

In the context of protein discovery, Alipanahi et al \cite{Alipanahi2015-qh} used sensitivity analysis to evaluate the relevance nucleotide mutations in the neural network prediction. The result is a mutation map that shows binding variations within a sequence. Vidovic \cite{Vidovic2016} proposed the Measure of Feature Importance (MFI), a metric that is intrinsically non-linear. It focuses on measuring how the interaction among features changes the prediction.   

As we have pointed out previously, Capsule Networks exhibit more intrinsic interpretability  properties, which is indeed a result of the routing-by-agreement algorithm. The instantiation parameter values of the activations vectors can be used to explain why the network detects certain features. Specifically, when all capsules of an object are in an appropriate relationship, the higher level capsule of that object should have a higher likelihood of activation. 

\section{Capsule Network Implementation}

\begin{table*}[ht]
\centering
\caption{Loss functions, optimizers, and hyperparameters that yield acceptable testing accuracy value in our experiments.}
\label{loss_opts_params}
\begin{tabular}{|c|c|c|}
\hline
\textbf{Dataset} & \textbf{Loss} & \textbf{Hyperparameters} \\\hline
2D KRAS-HRAS & categorical\_hinge & filters = 64, kernel\_size = 9 \\
             &                    & primarycap\_dim = 32, voxelcap\_dim = 64 \\\hline
3D KRAS-HRAS & categorical\_hinge & filters = 128, kernel\_size = 7 \\
             &                    & primarycap\_dim = 32, voxelcap\_dim = 64 \\\hline
2D PSI-BLAST & logcosh & filters = 512, kernel\_size = 5 \\
             &         & primarycap\_dim = 16, voxelcap\_dim = 32 \\\hline
3D PSI-BLAST & logcosh & filters = 64, kernel\_size = 5 \\
             &         & primarycap\_dim = 16, voxelcap\_dim = 32 \\\hline
\end{tabular}
\end{table*}

In this section, we discuss the proposed Capsule Network architecture along with the data representation for a protein classification problem.  

\subsection{DataSets}
The datasets used in this research are {\it KRAS-HRAS} and {\it PSI-BLAST}. The {\it RAS} family of proteins are of great interest in cancer research since these proteins are considered to be undruggable due to their lack of obvious cavities on their lobular surfaces \cite{McCormick2015-ji}. RAS family of proteins which are related to 95\%  of the pancreatic cancer and 45\% of colorectal cancer. The {\it KRAS-HRAS} dataset contains protein structures belonging to the {\it KRAS} and {\it HRAS} subfamilies of {\it RAS}. There is a total of 233 structures, 77 and 156 belonging to {\it KRAS} and {\it HRAS} respectively. The {\it PSI-BLAST} dataset contains protein structures obtained from a {\it PSI-BLAST} search using {\it RAS} sequences. This set was generated to test the ability of neural network to classify between {\it RAS} structures and structures that closely resemble {\it RAS}, but do not belong to the family. There is a total of 510 structures, 362 and 148 belonging to {\it RAS} and {\it Non-RAS}, respectively. 

\subsection{Data Representation}
We use the approach proposed by Corcoran et al. \cite{Corcoran2018-qb} to encode 3D  proteins from Protein Data Bank (PDB) files into 2D representations (voxel grid) by mapping a traversal of space-filling Hilbert curves. 3D PDB files are also easily transformed into 3D voxel cubes representations (no Hilbert curves mapping involved). These representations contain information of the atoms in the 3D protein structure, including the type of amino acid residue. These representations are the input data of our 2D and 3D Capsule Network implementations with a $512^2$ voxel grid (voxel size of 1{\AA}) and a $64^3$ voxel cube, (voxel size of 1{\AA}), respectively, and 8 channels of information. The 8 channel of information correspond to the following type of residues: aliphatic, aromatic, neutral, acid, basic, glycine, $\alpha$-carbon and $\beta$-carbon. 

\begin{figure}[htp]
\centering
\includegraphics[scale=1.5]{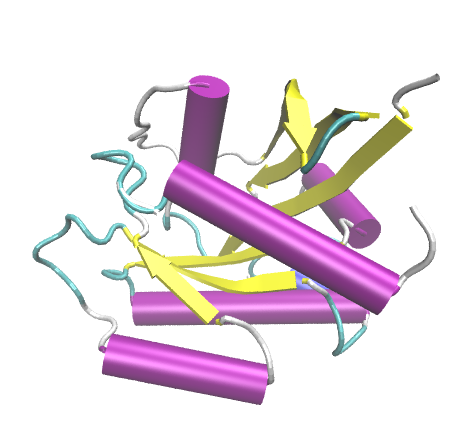}
\caption{New cartoon representation of the human K-Ras G12D Mutant in complex with GDP and Cyclic Inhibitory Peptide crystal structure (Rendered with VMD).}\label{5xco_newcartoon}
\end{figure}

\begin{figure}[htp]
\centering
\includegraphics[scale=0.14]{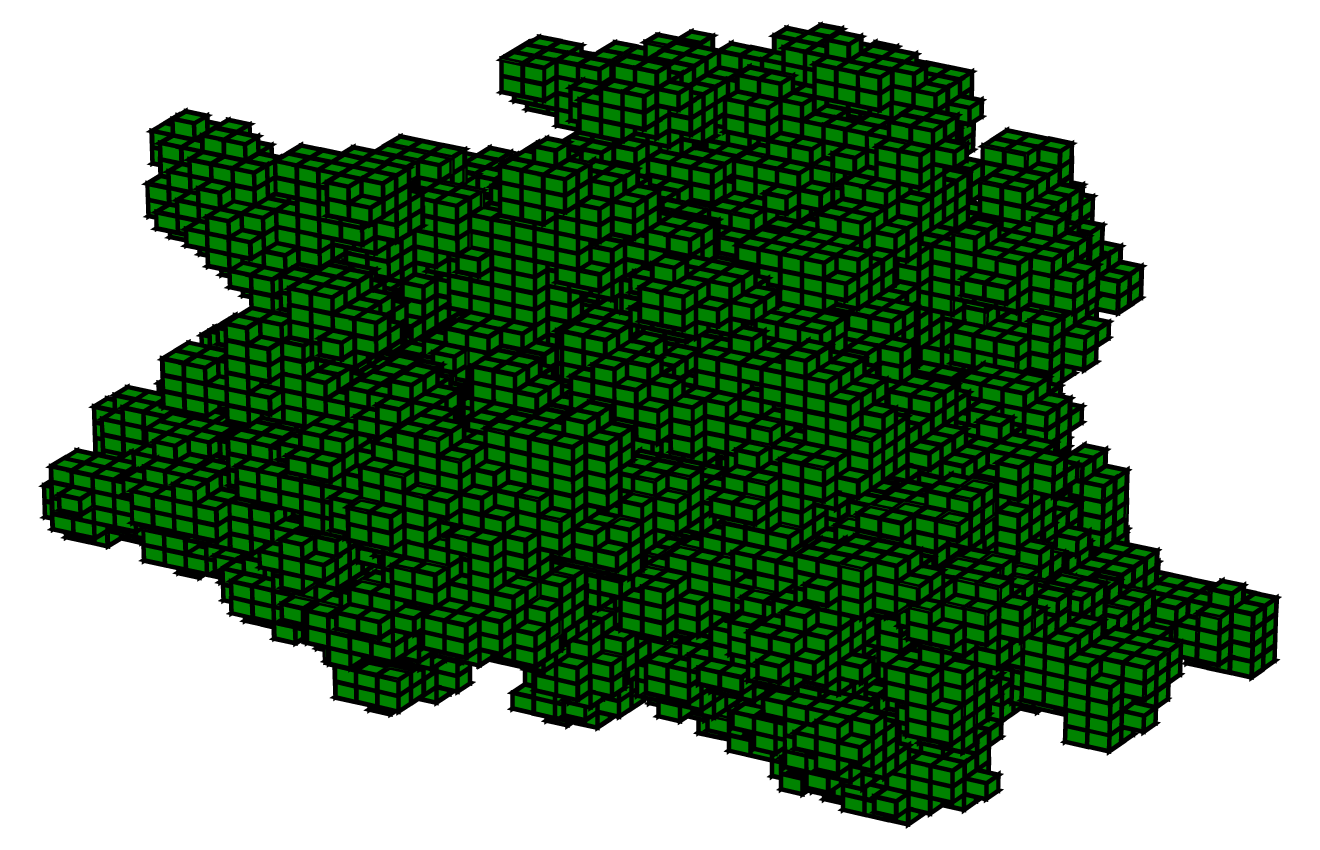}
\caption{3D voxelized representation.}\label{5xco_3dvoxels}
\end{figure}

\begin{figure}[htp]
\centering
\includegraphics[scale=0.1]{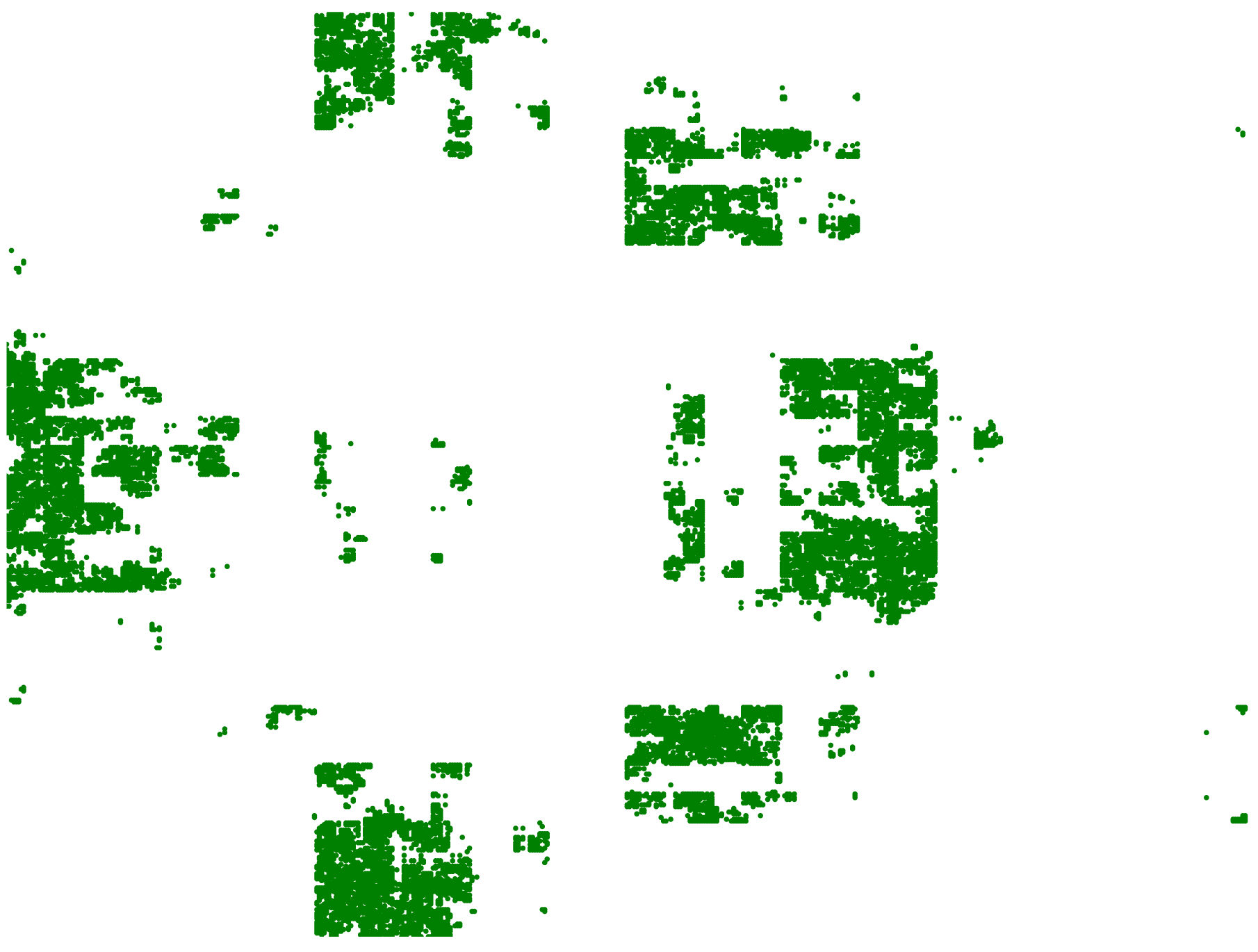}
\caption{2D voxelized filling curve representation.}\label{5xco_2dvoxels}
\end{figure}

Figures \ref{5xco_newcartoon} to \ref{5xco_2dvoxels} show a cartoon representation of a {\it RAS} protein instance along with its 3D voxel representation and the corresponding 2D mapping. 

\subsection{Capsule Network implementation}

\begin{figure}[htp]
\centering
\includegraphics[scale=0.3]{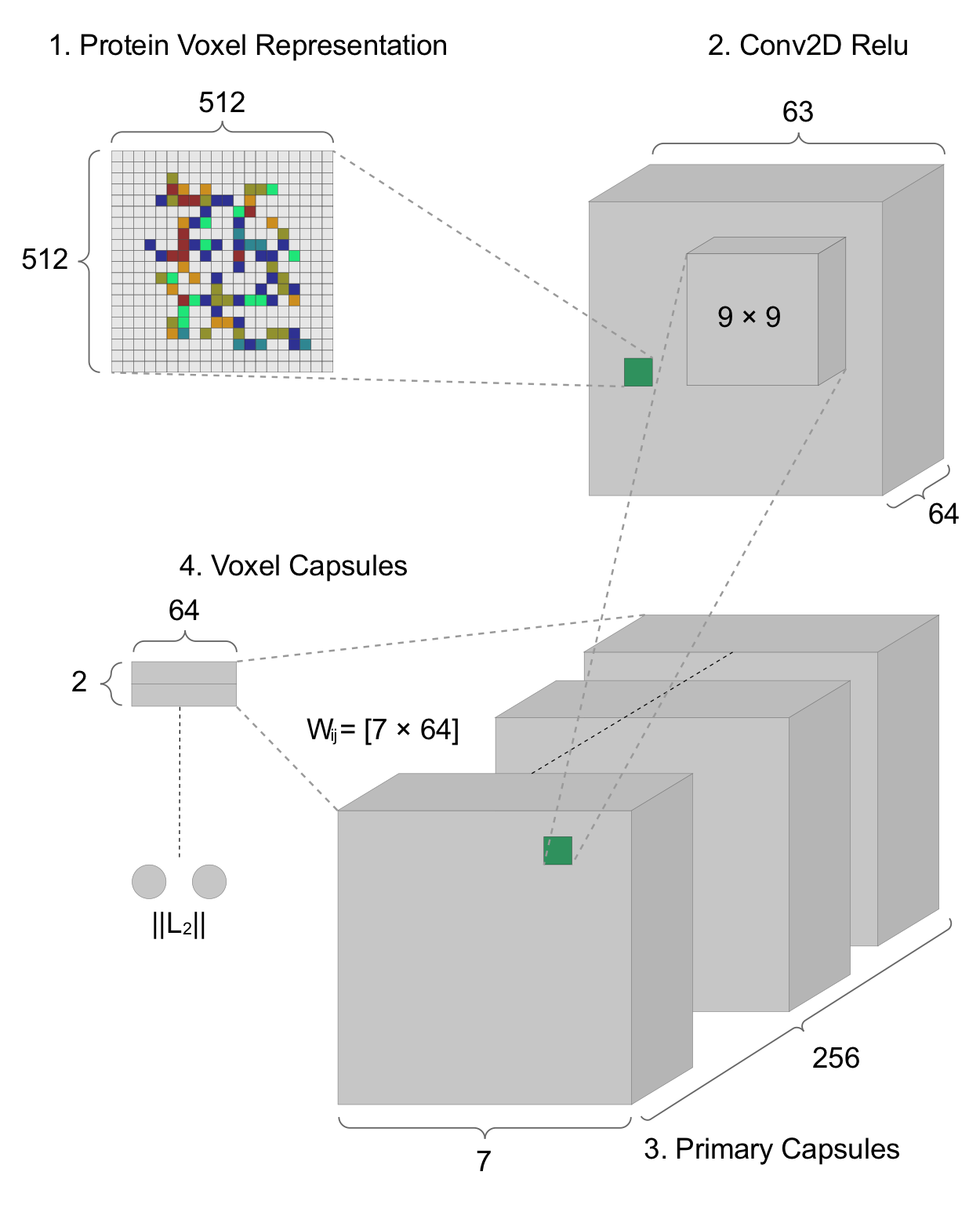}
\caption{Capsule Network Architecture}\label{archi}
\end{figure}

The proposed Capsule Network architecture is illustrated in Figure \ref{archi}. The input data is fed into a convolutional layer that detects basic features of the voxel representations. The output of this layer is passed to the primary capsule layer where a combination of the features detected is produced. In this layer the data is fed into a convolutional sub-layer and then passed to a reshape sub-layer that prepares the data for the squash operation before it is passed to the voxel capsule layer. In the voxel capsule layer, the dynamic routing operation occurs with 3 routing iterations. Finally, the data is passed to a length layer where each capsule in the voxel capsule layer is replaced with its length. This length represents the probability of a voxel capsule matching the label of a protein (i.e. {\it RAS} or {\it Non-RAS}).

Hinton's capsule Network architecture has a dense decoder at the end of the network for regularization purposes. In contrats, we do not use a decoder in our Capsule Network implementations due to scalability issues related to the size of our problem. Instead, we tested different combinations of loss functions, optimizers, and hyperparameters until our model reached acceptable results. The final selection of these functions and parameters are listed in Table \ref{loss_opts_params}. The categorical hinge \cite{Rosasco2004} function performs a summatory over the all the incorrect categories and compares the score of the correct and incorrect categories. If the score of the correct category is greater than the score of the incorrect one by some margin, the loss is 0. Otherwise, the loss is obtained by subtracting the score of the correct category from the incorrect one and adding it to 1. This makes the hinge function suitable for our classification problem since it is binomial. The \(log(cos(x))\) loss function is similar to the Huber loss function \cite{huber1964}, but it can be differentiated twice everywhere. It is approximately equals to \(\frac{x^2}{2}\) for small values of \(x\) and to \(|x| - log(2)\) for large values of \(x\) which avoid sensitivity to incorrect predictions.

\begin{table}[H]
\centering
\caption{RMSProp parameters.}
\label{rmsparams}
\begin{tabular}{|c|c|}
\hline
\textbf{Parameter} & \textbf{Value} \\\hline
Learning Rate & 0.001 \\\hline
Rho & 0.9 \\\hline
Epsilon & None \\\hline
Decay & 0.0\\\hline
\end{tabular}
\end{table}

We use Root Mean Squared Propagation (RMSProp) as  optimizer. RMSProp takes the sign of the last two gradients to increase or decrease the step size at which the decay is done by dividing the last gradient by the root mean squared of the moving average of the squared gradient for each weight. This is suitable for training, validating, and testing network models in batches as in our case where this was done batches of 1 sample. The values of the parameters for RMSProp are shown in Table \ref{rmsparams}. Five hyperparameters can be tuned including the dimensions of the primary and voxel capsules, number of filters, kernel size, and stride. The filters apply to the convolutional layer only. The kernel size applies to the convolutional and primary capsule layers while the primary and voxel  dimensions apply to the primary capsule layer and voxel capsule layer respectively. In terms of the stride, it was set to 8 for all the datasets.

\section{Results}

\subsection{Dataset 1: {\it KRAS-HRAS}}

For baseline performance, five convolutional neural networks \cite{Corcoran2018-qb} classify 2D representations of { \it KRAS } or { \it HRAS } proteins with a testing accuracy between 0.67 and 0.83. Our Capsule Network implementation obtained a testing accuracy of 0.94 for that same dataset. In this case, both training and validation reached values of more than 0.95 between 4 epochs and 8 epochs. The loss reached its minimum at the ninth epoch during both training and validation.

\begin{figure}[htp]
\centering
\includegraphics[scale=0.6]{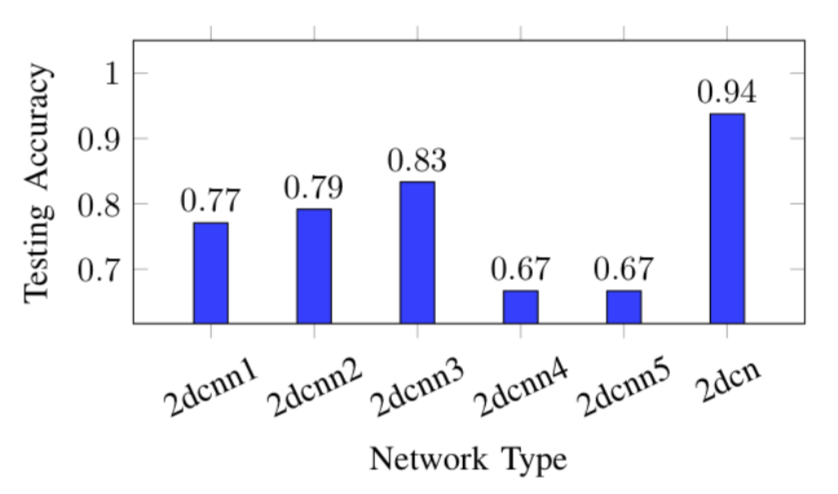}
\caption{Testing accuracy obtained by multiple neural network schemes on the 2D {\it KRAS-HRAS} dataset.}\label{KRAS-HRAS2d}
\end{figure}

Figure \ref{KRAS-HRAS2d} shows the accuracy results for the 2D case. similar results of accuracy are obtained in other cases.

In addition, 3 convolutional neural networks \cite{Corcoran2018-qb} classify 3D representations of { \it KRAS } or { \it HRAS } proteins with a testing accuracy between 0.65 and 0.77. Our Capsule Network implementation obtained a testing accuracy of 0.93 for that same dataset. The accuracy in training and validation reached values of more than 0.95 between 4 epochs and 8 epochs. The loss reached its minimum at the ninth epoch during both training and validation.

\subsection{Dataset 2: {\it PSI-BLAST} - RAS/Non-RAS}

Five convolutional neural networks \cite{Corcoran2018-qb} classify 2D representations of { \it RAS } or Not-{ \it RAS } proteins with a testing accuracy between 0.72 and 0.78. Our Capsule Network implementation obtained a testing accuracy of 0.87 for that same dataset. During training the network reached an accuracy of more than 0.95 after 3 epochs. However, during validation the accuracy reached a maximum of 0.88 and was steady between epochs 3 through 20 between 0.84 and 0.88. During training the loss began at 0.13 and stayed steady between 0.004 and 0.008 in epochs 10 through 20. During validation the loss began at 0.072 and stayed steady between 0.045 and 0.053 in epochs 3 through 20.

For the 3D case, 3 convolutional neural networks \cite{Corcoran2018-qb} classify 3D representations of { \it RAS } or { \it Non-RAS } proteins with a testing accuracy between 0.70 and 0.82. Our Capsule Network implementation obtained a testing accuracy of 0.85 for that same dataset. During training the network reached an accuracy of more than 0.95 after 3 epochs. However, during validation the accuracy reached a maximum of 0.92 and was steady between epochs 3 through 20 between 0.88 and 0.92. During training the loss began at 0.104 and stayed steady between 0.006 and 0.009 in epochs 12 through 20. During validation the loss began at 0.072 and stayed steady between 0.045 and 0.053 in epochs 3 through 20.

\begin{figure}[htp]
\centering
\includegraphics[scale=0.6]{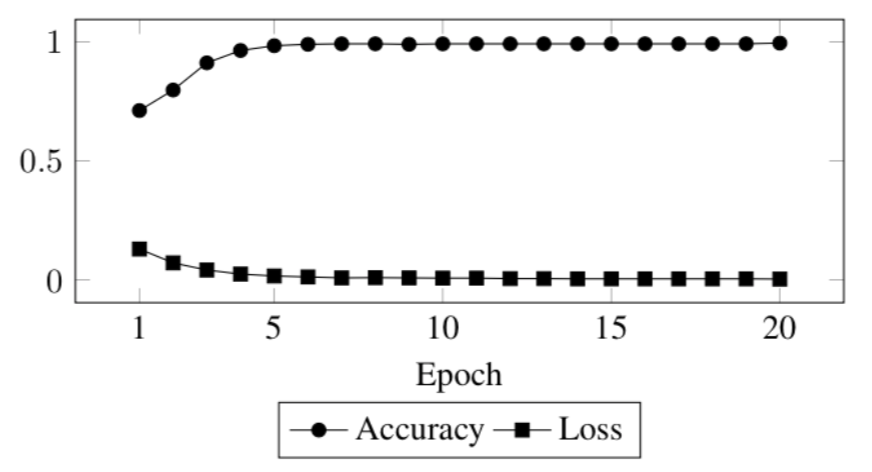}
\caption{Training accuracy and loss obtained by 2dcn on the 2D {\it PSI-BLAST} dataset in 20 epochs.}\label{psi-blast-2d-training}
\end{figure}

\begin{figure}[htp]
\centering
\includegraphics[scale=0.6]{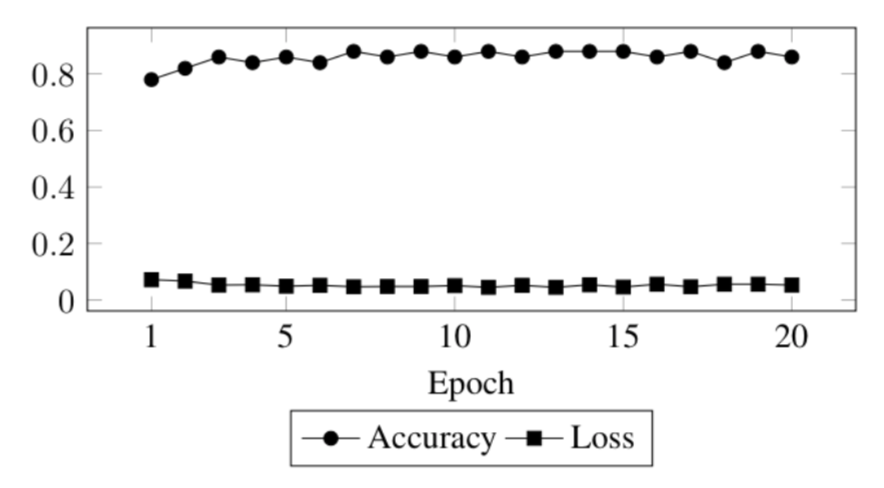}
\caption{Validation accuracy and loss obtained by 2dcn on the 2D {\it PSI-BLAST} dataset in 20 epochs.}\label{psi-blast-2d-validation}
\end{figure}

Accuracy and loss results are calculated in all cases. For example, Figures \ref{psi-blast-2d-training} and \ref{psi-blast-2d-validation} show the results obtained for the accuracy and loss during training and validation in 20 epochs for the {\it PSI-BLAST} 2D dataset.

\subsection{Runtime Performance}

Tables \ref{2d-khras} and \ref{3d-khras} summarizes the testing accuracy and running time obtained by multiple neural network schemes on the 2D and 3D {\it KRAS-HRAS} dataset with and without data augmentation. Similarly, Tables \ref{2d-psi-da} and \ref{3d-psi-da} summarizes the testing accuracy and running time results for the 2D and 3D {\it PSI-BLAST} dataset. It is worth to note that in all cases the Capsule Network implementation does not require data augmentation to reach high accuracy and consequently the computational time is decreased compared to the Convolutional Neural Network with data augmentation.

\begin{table}[H]
\centering
\caption{2D {\it KRAS-HRAS} dataset: Testing accuracy and running time}
\label{2d-khras}
\begin{tabular}{|c|c|c|}
\hline
\textbf{Network Type} & \textbf{Accuracy} & \textbf{Average Running Time (s)} \\\hline
CNN & 0.83 & 0.35\\\hline
CNN+DA & 0.84 & 14.5 \\\hline
CN & 0.94 & 4.78 \\\hline
CN+DA & 0.93 & 15.7 \\\hline
\end{tabular}
\end{table}

\begin{table}[H]
\centering
\caption{3D {\it KRAS-HRAS} dataset: Testing accuracy and running time.}\label{3d-khras}
\begin{tabular}{|c|c|c|}
\hline
\textbf{Network Type} & \textbf{Accuracy} & \textbf{Average Running Time (s)} \\\hline
CNN & 0.77 & 0.46 \\\hline
CNN+DA & 0.67 & 16.5 \\\hline
CN & 0.92 & 6.65 \\\hline
CN+DA & 0.92 & 17.9 \\\hline
\end{tabular}
\end{table}

\begin{table}[H]
\centering
\caption{ 2D {\it PSI-BLAST} dataset: Testing accuracy and running time.}
\label{2d-psi-da}
\begin{tabular}{|c|c|c|}
\hline
\textbf{Network Type} & \textbf{Accuracy} & \textbf{Average Running Time (s)} \\\hline
CNN & 0.78 & 0.68 \\\hline
CNN+DA & 0.83 & 28.0 \\\hline
CN & 0.87 & 1.98 \\\hline
CN+DA & 0.87 & 33.2 \\\hline
\end{tabular}
\end{table}

\begin{table}[H]
\centering
\caption{3D {\it PSI-BLAST} dataset: Testing accuracy and running time..}
\label{3d-psi-da}
\begin{tabular}{|c|c|c|}
\hline
\textbf{Network Type} & \textbf{Accuracy} & \textbf{Average Running Time (s)} \\\hline
CNN & 0.82 & 1.98 \\\hline
CNN+DA & 0.82 & 32.8 \\\hline
CN & 0.85 & 27.8 \\\hline
CN+DA & 0.85 & 36.4 \\\hline
\end{tabular}
\end{table}

\subsection{Protein Structure Prediction}

When predicting the classes of the samples in the two datasets, the accuracy of our model is calculated as follows:

\begin{equation}
\begin{aligned}
P_{acc} =  \frac{1}{|cp|}\sum_{i=1}^{|cp|} cp_i,\\
\forall cp_i \in [0, 1],
\end{aligned}
\end{equation}
where \(cp\) is a vector with values that indicate if the ith prediction is correct. The magnitude of \(cp\) is equal to the number of samples used during testing for each dataset. Our 2D and 3D Capsule Network implementation obtained a prediction accuracy of 0.92 in both 2D and 3D KRAS-HRAS datasets while for the 2D and 3D PSI-BLAST datasets they obtained 0.68 and 0.84 respectively.

\begin{figure}[H]
\centering
\includegraphics[scale=0.6]{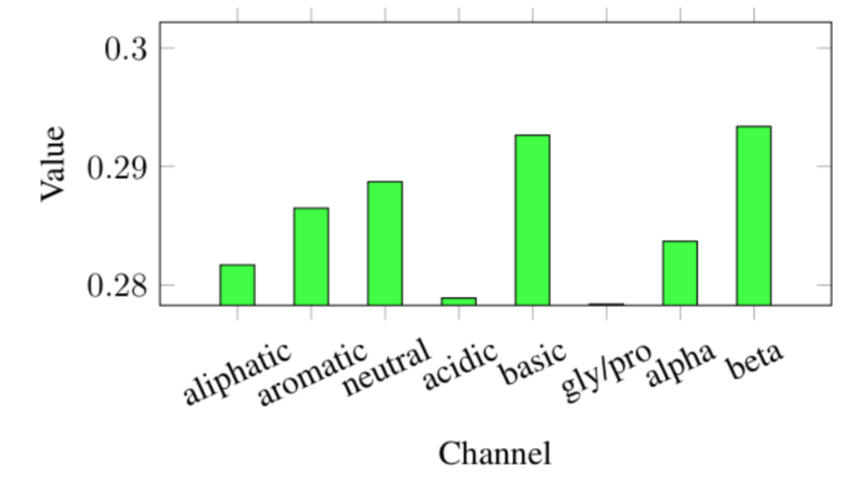}
\caption{Scores per channel for an instance of the {\it KRAS-HRAS} protein dataset}\label{krashras-scores-per-channel}
\end{figure}

Figure \ref{krashras-scores-per-channel} shows the scores per channel for an instance of the {\it KRAS-HRAS} dataset.
a high score values means high probability to be predicted correctly as part of the channels of information in the protein instance. 

\subsection{Interpretability}

To demonstrate the interpretability benefits of the Capsule Network implementation, we show how the activation vectors provide valuable information about the protein structure. After training the network, a specific PDB file is chosen and a modified version is created by changing information about some of its atoms (i.e. location in space, type of atom, etc.). Then, both the original and the modified versions are put through the network and their respective activation vectors are retrieved. Once obtained, the original output vectors are compared with the modified version’s by obtaining the distance between the individual elements and calculating the norm of the resulting vectors (one for each classification). This allows us to observe just how much each classification’s respective vector changes due to the modification made to the input, and therefore, analyze which parts of the input the network considers important for its classification.

\begin{figure}[H]
\centering
\includegraphics[scale=0.8]{5xco_one.png}
\includegraphics[scale=0.8]{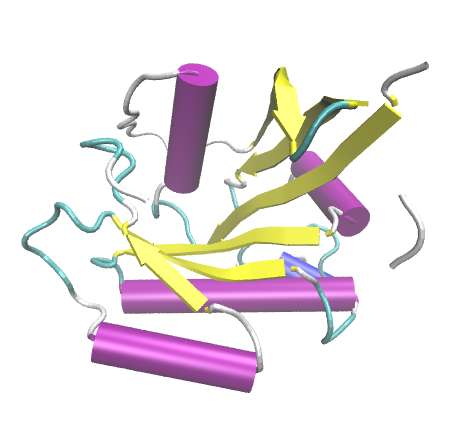}
\caption{Original and Modified 5XCO Protein cartoon }\label{originalpro}
\end{figure}

Figure \ref{originalpro} shows cartoon representations generated in VMD of the original protein, found in PDB file {\it 5XCO} of the KRAS-HRAS dataset, and the protein after removing the alpha helix with residue ids 152 through 166.

Table \ref{tableJC1} shows how the classification changes with regards to specific changes in the protein and how the changes are reflected into the most significant components of the activation vectors. 

\begin{table*}[ht]
\centering
\caption{Classification results from targeted changes, and the norm between the original activation vectors and the modified version vectors.}
\label{tableJC1}
\begin{tabular}{|c|c|c|c|c|}
\hline
\textbf{Change} & \textbf{Original} & \textbf{Modified} & \textbf{KRAS norm} & \textbf{HRAS norm} \\\hline
Removed Alpha Helix (87-104) &	HRAS	& HRAS	& 5.20E-02	& 1.25E-01\\\hline
Removed Alpha Helix (66-74)	& HRAS	& HRAS	& 0.17021	& 0.16162 \\\hline
Removed Alpha Helix (152-166) &	HRAS	& KRAS	& 1.0464	& 3.73E-03 \\\hline
Removed Coil (117-126)	& HRAS	& KRAS	& 1.0007	& 9.73E-01 \\\hline
Removed Coil (144 -151)	& HRAS	& HRAS	& 0.359	& 6.08E-01 \\\hline
\end{tabular}
\end{table*}

\subsection{Computational Resources}
The results were generated using the Keras 2.2.0 API \cite{keras} on top of the TensorFlow 1.8.8 \cite{Tensorflow} computation backend. We used nodes at Chameleon Cloud \cite{chameleloncloud} with 250GB of storage, 128GB of RAM, 48 Intel(R) Xeon(R) CPU E5-2670 v3 @ 2.30GHz, and two NVIDIA Tesla P100 GPU accelerators.

\section{Conclusion and Future Work}

In this work, we have discussed the implementation and application of a Capsule Network architecture to the classification of proteins. It shows the potential of Capsule Networks to tackle problems in structural biology. To the best of our knowledge, our team is the first one to apply Capsule Networks to this field. The results demonstrate a gain in accuracy of Capsule Networks when compared to traditional convolutional networks.

For future work, we plan to address the more complex problem of protein structure prediction using Capsule Networks.  A typical computational approach to protein structure prediction is to sample the protein conformational space using a large number of 3D structures known as decoys. The quality of these decoys is evaluated and the most optimal decoys are selected. The proper selection of decoys becomes an important factor for an accurate protein structure prediction. The selection is done through scoring functions that combine certain features to provide an indicator of decoy quality. However, current scoring functions do not consistently select the best decoys. Deep learning offers great potential to improve decoy scoring by using sets of annotated decoys and learning the relations between the features and decoy quality. Moreover, it provides an opportunity to determine other features that may produce betters scores.

\section*{Acknowledgment}

This work was supported in part by the U.S. Department of Energy under the Visiting Faculty Program (VFP), Berkeley Lab Undergraduate Faculty Fellowship (BLUFF) Program and the Sustainable Research Pathways (SRP) program. We thanks Dr. Xinlian Liu and Rafael Zamora-Resendiz for their feedback on this work. Computing allocations were provided through NSF funded Chameleon Cloud. 

\bibliographystyle{ACM-Reference-Format}
\bibliography{References}

\end{document}